\documentclass[letterpaper]{article} 
\usepackage[draft]{aaai2026}  
\usepackage{times}  
\usepackage{helvet}  
\usepackage{courier}  
\usepackage[hyphens]{url}  
\usepackage{graphicx} 
\usepackage{natbib}  
\usepackage{caption} 
\usepackage{amsmath}
\usepackage{amssymb}
\usepackage{mathtools}
\newcommand{\E}{\mathbb{E}}

\usepackage{algorithm}
\usepackage{algorithmic}
\usepackage{booktabs}
\usepackage{multirow}
\usepackage{graphicx}
\usepackage{pifont}
\usepackage{multirow}
\newcommand{\cmark}{\checkmark} 
\newcommand{\xmark}{\ding{55}}  
\usepackage{newfloat}
\usepackage{listings}

\usepackage{xcolor}
\usepackage{colortbl}

\DeclareCaptionStyle{ruled}{labelfont=normalfont,labelsep=colon,strut=off} 
\floatstyle{ruled}

\newfloat{listing}{tb}{lst}{}
\floatname{listing}{Listing}
\title{CTA-Flux: Integrating Chinese Cultural Semantics into High-Quality English Text-to-Image Communities}

\author {
    Yue Gong\textsuperscript{\rm 1 2 }\thanks{Equal contribution. 
    },
    Shanyuan Liu\textsuperscript{\rm 2 }\footnotemark[1],
    Liuzhuozheng Li\textsuperscript{\rm 2 }\footnotemark[1],
    Jian Zhu\textsuperscript{\rm 2 3},
    Bo Cheng\textsuperscript{\rm 2},
    Liebucha Wu\textsuperscript{\rm 2},
    Xiaoyu Wu\textsuperscript{\rm 2},
    Yuhang Ma\textsuperscript{\rm 2},
    Dawei Leng\textsuperscript{\rm 2}\thanks{ Corresponding author: Dawei Leng (lengdawei@360.cn).},
    Yuhui Yin\textsuperscript{\rm 2}
}
\affiliations {
    \textsuperscript{\rm 1}School of Computer Science and Engineering, Beihang University\\
    \textsuperscript{\rm 2}360 AI Research\\
    \textsuperscript{\rm 3} Nanjing University of Science and Technology\\
     yuegong@buaa.edu.cn, liushanyuan18@163.com, liuzhuozheng-li@outlook.com,lengdawei@360.cn
     
}

\begin{document}

\maketitle

\begin{abstract}

We proposed the Chinese Text Adapter-Flux (CTA-Flux). An adaptation method fits the Chinese text inputs to Flux---a powerful text-to-image (TTI) generative model initially trained on the English corpus. Despite the notable image generation ability conditioned on English text inputs, Flux performs poorly when processing non-English prompts, particularly due to linguistic and cultural biases inherent in predominantly English-centric training datasets. Existing approaches, such as translating non-English prompts into English or fine-tuning models for bilingual mappings, inadequately address culturally-specific semantics, compromising image authenticity and quality. To address this issue, we introduce a novel method to bridge Chinese semantic understanding with compatibility in English-centric TTI model communities. Existing approaches relying on ControlNet-like architectures typically require a massive parameter scale and lack direct control over Chinese semantics. In comparison, CTA-flux leverages MultiModal Diffusion Transformer (MMDiT) to control the Flux backbone directly, significantly reducing the number of parameters while enhancing the model's understanding of Chinese semantics. This integration significantly improves the generation quality and cultural authenticity without extensive retraining of the entire model, thus maintaining compatibility with existing text-to-image plugins such as LoRA, IP-Adapter, and ControlNet. Empirical evaluations demonstrate that CTA-flux supports Chinese and English prompts and achieves superior image generation quality, visual realism, and faithful depiction of Chinese semantics. 

\end{abstract}

\begin{figure*}[t]
  \centering
 \includegraphics[width=1.0\linewidth, height=0.23\linewidth]{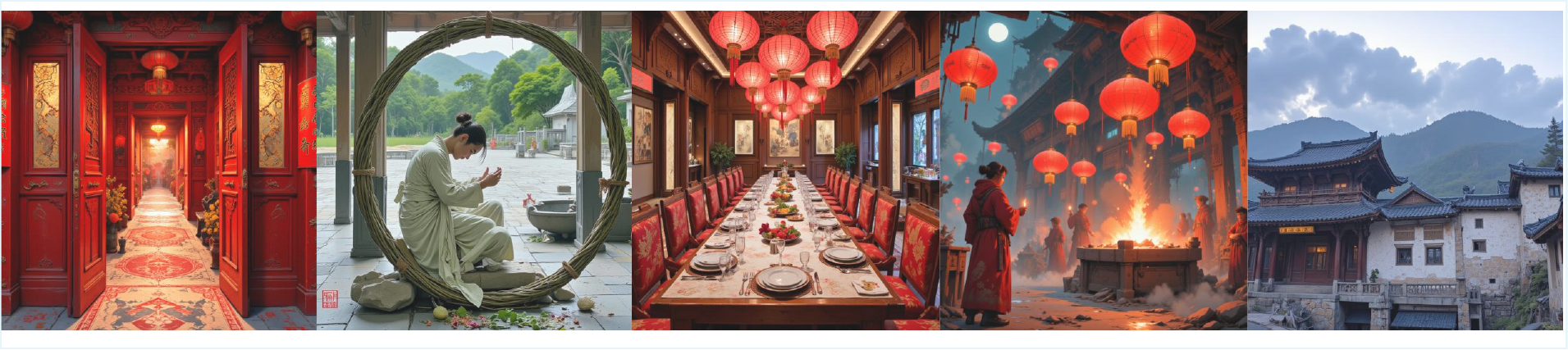} 
  \caption{Selected samples generated by our CTA-Flux model using Chinese text prompts.
 }
  \label{fig:temporal_vis}
\end{figure*}

\section{Introduction}
Recent advancements in text-to-image~(TTI) models have significantly enhanced the quality and diversity of generated images\cite{ramesh2021zero, saharia2022photorealistic, yu2022scaling}. The Stable Diffusion (SD) model\cite{rombach2022high, esser2024scaling} stands out among various TTI frameworks due to its remarkable capability in producing high-quality and photorealistic images from complex textual descriptions. SD's diffusion-based generative process~\cite{sohl2015deep,ho2020denoising,song2020score} gradually transports noisy input into clean images through iterative denoising steps, ensuring high fidelity and detailed visual outputs.
Concurrently, the emergence of the Flux framework\cite{flux2024} and its active community\cite{labs2025flux} has further boosted the development of TTI models. Flux's collaborative, open-source nature enables rapid evolution and community-driven improvements, leading to consistently improved image generation performance and better accessibility for diverse downstream applications.

However, despite these achievements, current TTI models, including those in the Flux community~\cite{labs2025flux}, still face limitations. A prominent challenge is that language-related bias originates from their training datasets, which are predominantly English-centric. Consequently, models often fail to perform well when generating images from non-English text prompts, struggling to capture the unique cultural semantics and nuanced symbols embedded in other languages~\cite{Liu_Cheng_Ma_Wu_Ma_Wu_Leng_Yin_2025}. This bias manifests primarily in two distribution gaps: the \textbf{linguistic feature distribution gap} and the \textbf{visual feature distribution gap}.

From the linguistic perspective, significant distribution gaps exist due to inherent language ambiguities and cultural differences. For instance, the English word "crane" can denote both a type of bird and a piece of heavy construction machinery, whereas the Chinese equivalent does not exhibit such polysemy. This results in semantic ambiguity when translating from languages that use words to express these concepts. Additionally, grammatical and structural disparities between languages further exacerbate the semantic mismatches and hinder the multilingual alignment\cite{yang2025qwen3}.

Beyond linguistic factors, the visual feature distribution gap—driven by cultural biases and implicit symbolic meanings in non-English text input—poses additional challenges. For example, when describing a person's appearance, Chinese prompts typically result in imagery featuring Asian characteristics and culturally specific contexts. In contrast, equivalent English descriptions predominantly yield imagery reflective of Western appearances and environments. Such implicit cultural distributions and symbolic representations are challenging to capture through direct translation or simple fine-tuning approaches.

Existing approaches attempt to address these challenges but remain inadequate under complex semantic conditioning. Straightforward approaches, such as translating non-English prompts into English before image generation\cite{fengshenbang}, are inadequate for TTI generation under complex semantic conditioning. These translation-based approaches lose culturally-specific meanings, subtle connotations, and symbolic expressions that are hard to translate, thus directly deteriorating image quality and authenticity. Another strategy involves fine-tuning existing models to learn the mappings between non-English and English textual representations\cite{chen2022altclip}, aiming to improve their understanding of non-English prompts. Nevertheless, such fine-tuned models often struggle with culture-specific semantics and nuanced concepts unique to the source language, as these meanings frequently lack concise and accurate equivalents in English. 
More radical approaches involve training entirely new models from scratch to address multilingual contexts \cite {feng2023ernie, gu2022wukong}. This requires massive computational resources and long-term training, making it prohibitive to most users. Furthermore, retraining often compromises compatibility with existing plugins and extensions developed by existing community-driven frameworks.

To address the challenges of multilingual TTI models, we propose a novel approach that integrates a multilingual adaptation branch into the existing model architecture, ensuring strong compatibility with the Flux community framework. Specifically, our method incorporates non-English embedding tokens into the cross-attention mechanism of the MultiModal Diffusion Transformer~(MMDiT) model by a language adapter. This integration allows non-English linguistic control information to directly influence the backbone while maintaining compatibility with existing community models, enabling the model to handle non-English prompts without altering the integrity of the original architecture. Our approach introduces minimal additional parameters, maintaining a lightweight design that ensures efficiency and scalability. At the same time, the backbone model remains fixed, allowing the method to be easily combined with other Flux community plugins.

Moreover, to effectively address the semantic and cultural differences between languages, we employ a sophisticated two-stage training strategy. The model is trained using English and non-English data in the first stage. During this phase, we use a \textbf{Representation Alignment Loss} to align the features extracted by the non-English language encoder with the feature space of the original English language encoder. This alignment minimizes the linguistic feature domain gap and prevents the multilingual branch from needing to learn basic semantic concepts from scratch, accelerating the learning process. In the second stage, the model is fine-tuned exclusively on non-English data that contains culturally specific concepts. This allows the model to learn unique visual distributions associated with different cultures, enabling it to generate culturally accurate images and be visually faithful to the non-English prompts. Through this two-stage training, our approach accelerates convergence while enabling the model to capture both universal cross-lingual concepts and culture-specific semantics, thereby improving the quality and accuracy of generated images.

Our approach offers several key advantages: due to the lightweight design of the language adapter modules, it introduces minimal additional parameters to the backbone model. Additionally, it preserves the original Flux model backbone parameters, ensuring seamless integration with existing Flux community plugins and tools. Notably, our method also enhances the model's ability to generate images that capture unique cultural and linguistic symbols, which helps mitigate data bias and discrimination, especially for non-English languages. This ensures more inclusive and accurate image generation, reducing the risks of biased or culturally inappropriate outputs.

To summarize, our work contributes:

\begin{itemize}
    \item We propose a novel method for natively adapting the Flux framework to non-English languages, specifically focusing on \textbf{Chinese} language support. To the best of our knowledge, this is the first work to adapt Flux to handle Chinese language prompts effectively.
    \item Our approach not only extends the capabilities of Flux to non-English environments such as Chinese but also ensures seamless compatibility with existing plugins developed within the English-speaking Flux community. This multilingual compatibility makes our method versatile and capable of bridging the gap between English-centric and Chinese use cases.
    \item Experimental results demonstrate that our method performs consistently well across various metrics and benchmarks, including both general image generation tasks and those requiring cultural specificity, validating its effectiveness in multilingual generation settings.
\end{itemize}

\section{Related Work}
\subsection{Diffusion Models and Flow Matching}

Diffusion models (DMs)~\cite{ho2020denoising, sohl2015deep} are powerful generative models that gradually generate images from a random Gaussian noise with a deep denoising model. Score-based generative models (SGMs)~\cite{song2019generative,song2020score} learn to reverse a forward Ito process using stochastic differential equations (SDEs), training a score function $\nabla_{\boldsymbol{x}} \log p(\boldsymbol{x}, t)$ via denoising score matching (DSM) to guide sampling. Techniques like DDIM~\cite{song2020denoising}, ADM~\cite{dhariwal2021diffusion}, CM~\cite{song2023consistency}, and latent diffusion~\cite{rombach2022high} improve efficiency and output quality. Flow matching (FM)~\cite{lipman2023flow} and rectified flow~\cite{liu2022flow} offer deterministic alternatives by learning velocity fields over ODEs, enabling few-step or even single-step generation~\cite{yin2024one}. FM and SGMs are theoretically linked through Fokker-Planck equations and probability flow ODEs~\cite{song2020score,evans2010partial}, inspiring hybrid models that balance diffusion’s robustness with flow’s efficiency~\cite{esser2024scaling,flux2024, labs2025flux}.

\subsection{On the adaptation of diffusion generative models}
Recent advances in generative foundation models have facilitated the generation of task-specific outputs without the need to train large backbone networks from scratch. A variety of adaptation methods have been proposed to support this capability. IP-Adapter~\cite{ye2023ipadaptertextcompatibleimage} adds lightweight cross-attention modules to inject reference images as prompts, enabling efficient style transfer and image-guided generation. ControlNet~\cite{zhang2023adding} attaches task-specific control branches (e.g., depth, pose, edges) to frozen DMs for precise structural control. Other parameter-efficient methods like LoRA and textual inversion~\cite{hu2022lora,gal2022imageworthwordpersonalizing} reduce compute costs while maintaining quality. In flow matching (FM), models such as Flux. 1 and Flux Kontext~\cite{flux2024} use deterministic flows for fast generation and editing, integrating well with adapter-based control, suggesting unified controllability across DMs and FM frameworks.

Multilingual adaptation is important for T2I generative models as it enables global accessibility, allowing users to use the model in their native languages. While previous works have made progress in supporting multilingual input, they often treated non-English languages as auxiliary. For example, \cite{Li_2023} mitigated the language gap by translating English captions into other languages using a neural machine translation system. AltCLIP \cite{chen2022altclip} took a different route by enhancing diffusion models with the multilingual text encoder XLM-R. Meanwhile, ERNIE-ViLG 2.0 \cite{feng2023ernievilg20improvingtexttoimage} trained a diffusion model using Chinese image-text pairs from scratch. However, these approaches did not effectively bridge native language and English-speaking communities. \cite{Liu_Cheng_Ma_Wu_Ma_Wu_Leng_Yin_2025} proposed the bridge diffusion, but it can only be applied to Unet-based models such as Stable Diffusion 1.5. In contrast, our work integrates native language models into the pretrained English-centric DiT, aiming to achieve a state-of-the-art multilingual generation framework.

\section{Preliminary}

We denote $\boldsymbol{x}$ as the latent image embedding obtained from the VAE encoder. Let $\boldsymbol{\tau}_{EN}$ and $\boldsymbol{\tau}_{CN}$ represent the English and Chinese text embeddings $\boldsymbol{y}_{EN},\ \boldsymbol{y}_{CN}$  after the \textbf{T5} \cite{ni2021sentencet5scalablesentenceencoders} and \textbf{Qwen} \cite{yang2025qwen3} text encoders, respectively. We aim to train a model that generates images conditioned on a Chinese or an English text prompt. Formally, we strive to approximate the conditional distribution $p(\boldsymbol{x} \mid \boldsymbol{y})$, where we use $\boldsymbol{y}$ as a simplified notation for the text condition.

Under given text conditions $\boldsymbol{y}$, flow models, such as flux.1 \cite{flux2024} utilize a velocity field $\boldsymbol{v}(\boldsymbol{x},\boldsymbol{y}, t)$ to gradually turn noise $\boldsymbol{x}_{0},\boldsymbol{\epsilon}\in\mathcal{N}(\boldsymbol{0}, \boldsymbol{I})$ at $t=0$ into the data distribution $\boldsymbol{x}_1 \in p_{data}(\boldsymbol{x}|\boldsymbol{y})$ at $t=1$. The dynamics of the marginal distribution $p(\boldsymbol{x}, t|y)$ are formulated by probability flow ODE and the continuity equations:
\begin{equation}
    \left \{ 
    \begin{aligned}
        &\frac{\partial}{\partial t} p(\boldsymbol{x}, t|\boldsymbol{y}) = -\nabla_{\boldsymbol{x}} \cdot (\boldsymbol{v}(\boldsymbol{x},\boldsymbol{y}, t) \cdot p(\boldsymbol{x}, t|\boldsymbol{y})), \\
        &\boldsymbol{x}_0 \sim p_0(\boldsymbol{x}|\boldsymbol{y}) = \mathcal{N}(\boldsymbol{0}, \boldsymbol{I}), \quad \boldsymbol{x}_1 \sim p_{data}(\boldsymbol{x}|\boldsymbol{y}).
    \end{aligned}
    \right.
    \label{eq_continue_constraint}
\end{equation}
In application, our adaptation of Flux to Chinese input follows a similar optimization strategy, with the primary difference being that the English embeddings $\boldsymbol{\tau}_{EN}$ are replaced by the Chinese embeddings $\boldsymbol{\tau}_{CN}$. To estimate the velocity field conditioned on a Chinese text prompt, we train a diffusion transformer to estimate the velocity field, denoted as $\boldsymbol{v_{\theta}}(\boldsymbol{x}, \boldsymbol{y},t)$, using a dataset consisting of images and Chinese captions: $\{\mathcal{X}_{train}, \mathcal{Y}_{train}\}$, via the flow matching loss:
\begin{equation}
    \left \{ 
    \begin{aligned}
            \mathcal{L}_{\boldsymbol{\theta}} &= \E_{t, \boldsymbol{x}_i, \boldsymbol{\tau}^i_{CN}} \left[\left\| \boldsymbol{v_{\theta}}(\boldsymbol{x},\boldsymbol{\tau}^i_{CN} ,t)- (\boldsymbol{x}_i-\boldsymbol{\epsilon}) \right\|_2^2\right] \\
        \boldsymbol{x} &= (1-t)\cdot\boldsymbol{x}_i+t\boldsymbol\cdot \boldsymbol{\epsilon},
    \end{aligned}
    \right.
    \label{eq_loss}
\end{equation}
where $t\sim \mathcal{U}(0, 1), \  \boldsymbol{x}_i\sim \mathcal{X}_{train}$, $\boldsymbol{y}^{i}_{CN} \sim \mathcal{Y}^{CN}_{train}$. The velocity field $\boldsymbol{v_{\theta}}$ is estimated with a transformer backbone such as DiT. During the training and inference, the English and Chinese text prompts are passed through the text encoder to obtain the text embeddings, denoted as $\boldsymbol{\tau}_{EN}$ and $\boldsymbol{\tau}_{CN}$. 

\section{Method}

\begin{figure*}[t]
  \centering
  \includegraphics[width=0.74\linewidth, height=0.27\linewidth]{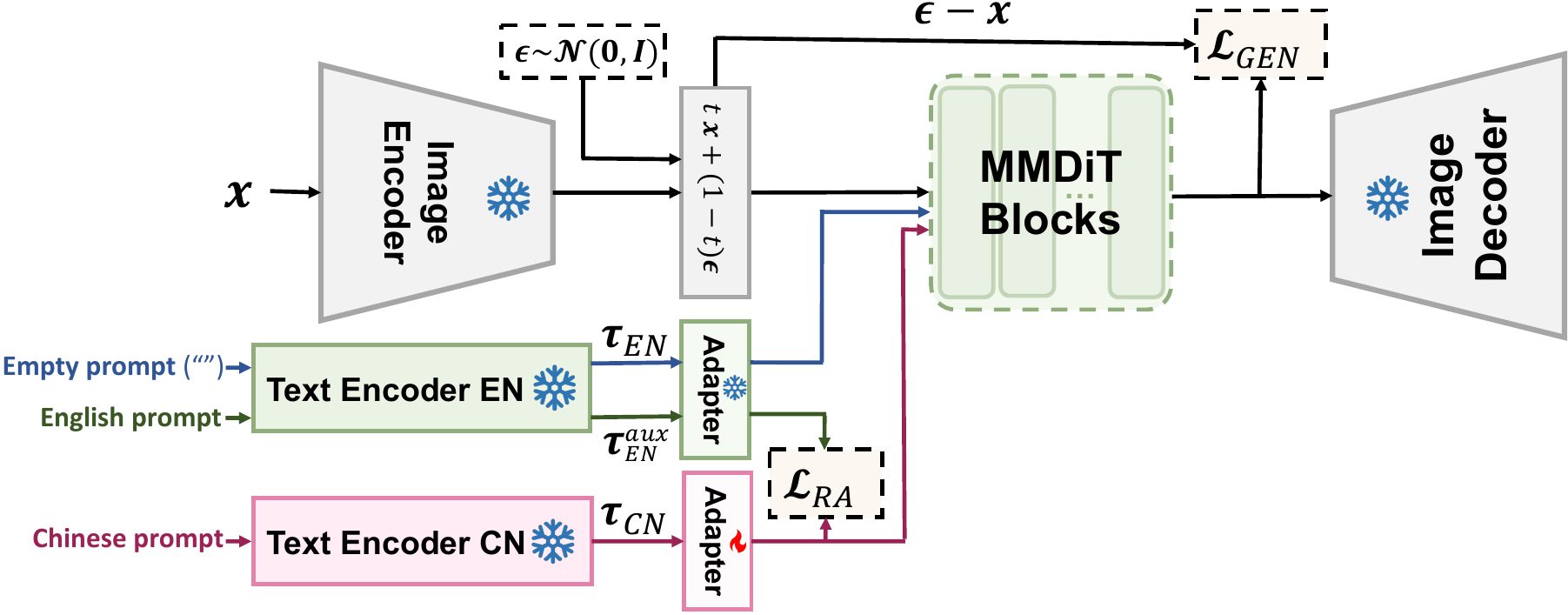} 
  \caption{The overall pipeline of our proposed CTA-Flux. The parameters of the image/text encoder and decoder are frozen during the training, and we optimize the Chinese project layer in MMDiT blocks. The model is optimized with the flow matching generation loss (denoted as $\mathcal{L}_{GEN}$) and representation alignment loss (denoted as $\mathcal{L}_{RA}$).
 }
  \label{fig:global_pipeline}
   \includegraphics[ height=0.39\linewidth]{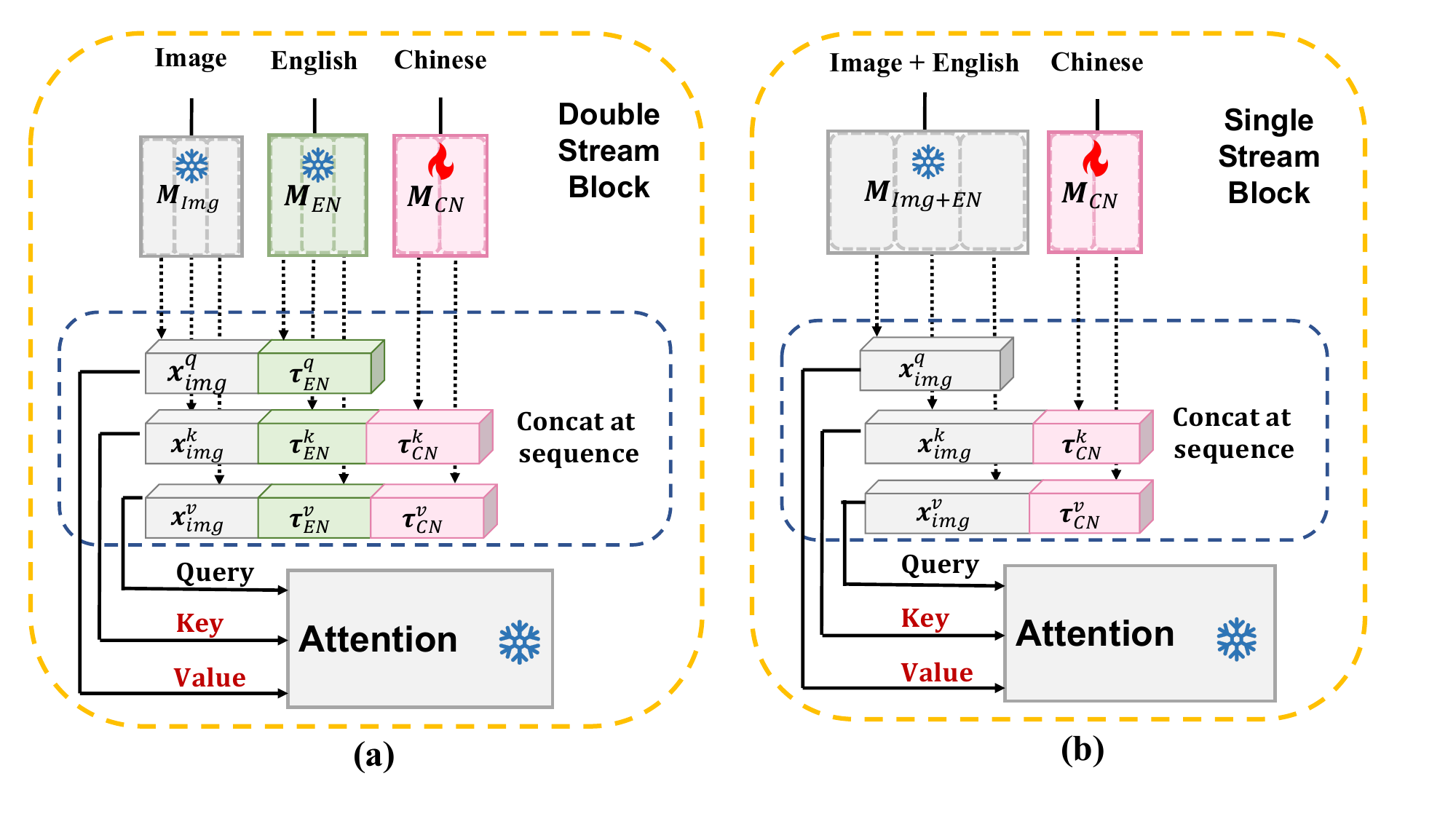}
  \caption{The frameworks of our Chinese Linguistic Attention Branch (CLAB) in the MMDiT module, each module consists of a double stream block and a single stream block. (a) and (b) shows the modification of the single stream and double stream block, respectively, where we add linear projection modules to map the Chinese embedding to Query, Key, and Value for the attention processor, denoted as $\boldsymbol{M}_{CN}$. 
 }
 \label{fig:MMDIT_cropped}
  \vskip -0.1in
\end{figure*}

Our overall framework, illustrated in Figure \ref{fig:global_pipeline}, consists of two primary components: the visual backbone and the language branch. To maintain compatibility with community-developed plugins built on base generative models, we adopt the pretrained Flux diffusion transformer as our backbone. In addition, we employ the same pretrained pixel Encoder and Decoder modules as those used in Flux to provide image latent embeddings. On top of this foundation, we design a parallel language branch dedicated to processing Chinese text prompts. This branch delivers Chinese linguistic information, functioning either in conjunction with English inputs or on its own, to enhance the model’s multilingual understanding and generation capabilities.

\subsection{The Architecture of the Language Branch}

As illustrated in Figure \ref{fig:global_pipeline}, the multilingual text inputs are passed through the Chinese Linguistic Attention Branch (CLAB) \cite{vaswani2023attentionneed}. Specifically, Chinese textual prompts are initially processed by a language-specific encoder capable of effectively capturing Chinese semantics. In our framework, we adopt \textbf{Qwen2.5}\cite{yang2025qwen3} as the encoder for Chinese language inputs, while retaining \textbf{T5}\cite{ni2021sentencet5scalablesentenceencoders} as the encoder for English inputs, thereby providing language embedding tokens for conditional generation. These embeddings subsequently undergo dimensional alignment with the backbone branch's latent dimensions via a multi-layer perceptron~(MLP)(Represented as the "English Adapter" and "Chinese Adapter" in Figure \ref{fig:global_pipeline}).

Given the intrinsic modality discrepancies between linguistic and visual domains, we employ a cross-attention mechanism rather than simple concatenation to integrate language information into the visual latents to facilitate the generation process effectively. As illustrated in Figure \ref{fig:MMDIT_cropped}, the aligned language tokens are projected into queries (conditionally updated based on task requirements), keys, and values (QKV) via dedicated projection layers at each layer within the MMDiT block of the backbone branch. The modified MMDiT blocks consist of a double-stream block (Figure \ref{fig:MMDIT_cropped} (a)) and a single-stream block (Figure \ref{fig:MMDIT_cropped} (b)). For the \textbf{double-stream block}, the Image and English tokens are passed through their respective frozen QKV projection layers ($\boldsymbol{M}_{img}$ and $\boldsymbol{M}_{EN}$) to generate QKV representations (denoted as $\boldsymbol{x}_{img}^q, \boldsymbol{x}_{img}^k, \boldsymbol{x}_{img}^v$ and $\boldsymbol{\tau}_{EN}^q, \boldsymbol{\tau}_{EN}^k, \boldsymbol{\tau}_{EN}^v$) for the attention processor. In the \textbf{single-stream block}, the Image and English input tokens are first concatenated and then passed through a frozen shared projection layer ($\boldsymbol{M}_{img+EN}$) to produce mixed QKV representations (denoted as $\boldsymbol{x}_{img+EN}^q, \boldsymbol{x}_{img+EN}^k, \boldsymbol{x}_{img+EN}^v$). For both the double-stream and single-stream blocks, the Chinese text embeddings are processed separately using a \textbf{trainable} Chinese projection layer $\boldsymbol{M}_{CN}$ to generate their corresponding KV embeddings: $\boldsymbol{\tau}_{CN}^k, \boldsymbol{\tau}_{CN}^v$.
These projected Chinese language embeddings interact with their image/English pairs in the MMDiT blocks through cross-attention operations.

This cross-attention strategy enables explicit and dynamic integration of non-English linguistic semantics into the backbone branch. Consequently, the CLAB provides nuanced, culturally-aware control signals during image synthesis, significantly enhancing the generation quality and semantic fidelity of outputs conditioned on non-English prompts. To ensure more effective training of the Language Branch and enable the model to learn control signals from non-English inputs accurately, we feed an empty string (“”) as the English text inputs to the English language branch during training. This strategy prevents the model from relying on the pretrained Backbone’s inherent ability to understand English prompts, forcing it to attend to the information provided by the Chinese language branch.

It is important to note that the English embedding projection layers in the Language Branch are frozen and not updated across the MMDiT blocks, and the non-English tokens are not used as queries in the cross-attention computation. This design stems from our observation that when Chinese branch tokens are used as queries in cross-attention, they inevitably attend to features from the English empty-text tokens in the Backbone Branch, which disrupts the Chinese control signals and results in semantically unstable or even random generations. By discarding the Chinese query tokens to prevent such interference, we maintain the effectiveness of Chinese text prompt conditioning throughout the generation process.

\subsection{Adaptation Strategy}

Our method preserves the architecture and parameters of the backbone branch entirely to ensure compatibility with existing community-developed plugins—such as ControlNet, LoRA, and custom checkpoints. Only the Chinese language branch parameters are updated during training, allowing our model to remain compatible with widely used tools and extensions within the Flux ecosystem.

Cross-lingual image generation presents two significant challenges: the linguistic feature distribution gap and  visual feature distribution gap. The former arises from inherent linguistic differences across languages, and the latter stems from cultural biases embedded in the training data. To address these issues, we propose a carefully designed two-stage adaptation strategy. In the first stage,  we mitigate the feature domain gap between the English and non-English language encoders by aligning their embedding spaces, enabling the Language Branch to capture shared semantics more effectively. In the second stage, we fine-tune the model exclusively on non-English data containing culturally specific concepts, thereby enabling the model to adapt to the target distribution shifts and generate images that reflect the unique cultural semantics of the source language.

In the first stage of training, we use a mixture of Chinese and English prompt texts. Each prompt—whether in Chinese or English—is passed through the Language Branch. We introduce an auxiliary supervision signal in addition to the standard image generation loss to accelerate the learning process and address the cross-lingual adaptation challenge. This auxiliary objective mitigates the “linguistic feature distribution gap” between non-English and English encoders.

Concretely, we use each non-English prompt’s semantically equivalent English counterpart solely to obtain features from the original English text encoder in the Backbone Branch. These English features are projected by the fixed MLP projection layer in the DiT backbone to match the DiT latent dimension, and are used only for computing the auxiliary alignment loss, without participating in the image generation process. Simultaneously, we obtain the projected non-English features from the Language Branch and minimize the mean squared error (MSE) between the projected non-English and corresponding English features. This auxiliary loss encourages the Language Branch to produce features compatible with the backbone’s embedding space and facilitates effective multilingual adaptation. The formulation of this loss is given by:
\begin{equation}
\left \{
\begin{aligned}
    &\mathcal{L}_{p} = MSE\left(\text{AvgPool}(\tau_{CN}), \text{AvgP}(\tau_{EN}^{aux})\right), \\
    &\mathcal{L}_{inter} = MSE\left(\tau_{CN}, \text{Intp}(\tau_{EN}^{aux}, \text{len}(\tau_{CN}))\right), \\
\end{aligned}
\right.
\end{equation}
where \(\tau_{EN}^{aux}\) represents the text embeddings derived from an English prompt semantically equivalent to the Chinese prompt employed to generate \(\tau_{CN}\), obtained via English the text encoder, $MSE(\cdot)$ denotes the mean square error between two tensors, $\text{AvgP}(\cdot)$ denotes the average pooling on the sequence dimension of tokens, and Intp($\cdot$) interpolates the English feature to the length of the Chinese feature in the sequence dimension. The loss function consists of two parts: $\mathcal{L}_p$   and $\mathcal{L}_{inter}$. $\mathcal{L}_p$ is the Mean Squared Error (MSE) calculated on the feature after pooling the entire sentence, representing the overall alignment of the sentence-level representation. $\mathcal{L}_{inter}$ is the MSE between the text encoder's English tokens, which are interpolated to match the length of the Chinese tokens in the text encoder, aiming to align the features across the different token lengths.

In the first stage of training, we found that incorporating a threshold $\mathcal{D}_{threshold}$ for the auxiliary loss is necessary. If the auxiliary loss falls below this threshold, it is set to zero. The purpose of this approach is to avoid overemphasizing alignment, allowing the model to retain the subtle differences inherent in the language itself. This ensures that there is greater optimization potential for improving the generation quality. The alignment loss is formulated as:
\begin{equation}
        \mathcal{L}_{RA} = \begin{cases} 
    \mathcal{L}_{p} + L_{inter} & \text{if } \mathcal{L}_{RA} \geq \mathcal{D}_{threshold} \\
    \mathcal{L}_{p} & \text{if } \mathcal{L}_{RA} < \mathcal{D}_{threshold}
    .
    \end{cases}
\end{equation}

In the second stage of training, we use solely Chinese data and incorporate a wide range of concepts unique to Chinese culture, such as special holidays, cuisine, and traditional clothing. The training resolution is gradually increased, starting from 256, progressing to 512, and finally reaching 1024, to enable the model to capture culturally specific output targets. The overall training objective function $\mathcal{L}_{\boldsymbol{\theta}}$ is as follows:
\begin{equation}
    \mathcal{L}_{\boldsymbol{\theta}} = \mathcal{L}_{GEN}+\mathcal{L}_{RA},
\end{equation}
where $\mathcal{L}_{GEN}$ follows the formulation in Eq.~\ref{eq_loss}, and $\mathcal{L}_{RA}$ denotes the alignment loss.

\subsection{Inference Strategy}

We can maintain consistency with the training phase during the inference phase by leaving the backbone text empty. Alternatively, we can input the same meaning in English into the backbone while inputting Chinese into the branch. We found that when Chinese is input into the branch while English is input into the backbone, it achieves better results than either using only the backbone or the branch independently.

\section{Experiments}
\subsection{Experimental Setup}
The training dataset consists of approximately one billion image-text pairs, primarily drawn from high-quality internal Chinese datasets, supplemented with publicly available English datasets. Roughly 60\% of the data is Chinese and 40\% is English. We ensure that Chinese data constitutes the majority to enhance the model’s capacity to generate semantically rich Chinese content and reduce the dominance of English-centric concepts. Before training, rigorous filtering is applied to exclude samples with watermarks, low aesthetic quality, or weak image-text alignment
We adopt FLUX.1-dev~\cite{flux2024} as the base generative model. For the Chinese text encoder, we use Qwen2.5-VL-7B-Instruct~\cite{bai2025qwen25vltechnicalreport}, and for the English text encoder, we adopt T5~\cite{ni2021sentencet5scalablesentenceencoders}, with the framework consistent with those used in FLUX.1-dev, and the weights of the T5 text encoder are frozen during the training. During training, the hyperparameter $\mathcal{D}_{\text{threshold}}$ is set to 0.05.

The training of the diffusion transformer is built upon Flux's latent space, which employs a Variational Autoencoder (VAE) to transform images between pixel space and latent representations. The training pipeline is implemented in PyTorch-Diffusers \cite{von-platen-etal-2022-diffusers}. We use the AdamW optimizer \cite{kingma2017adammethodstochasticoptimization, loshchilov2019decoupledweightdecayregularization} with a learning rate of 1e-5 and a total batch size of 3200. Training is conducted over two weeks on 32 NVIDIA A800 GPUs.

\subsection{Training Datasets}

\begin{figure}[ht]
  \centering
  \includegraphics[width=0.6\linewidth]{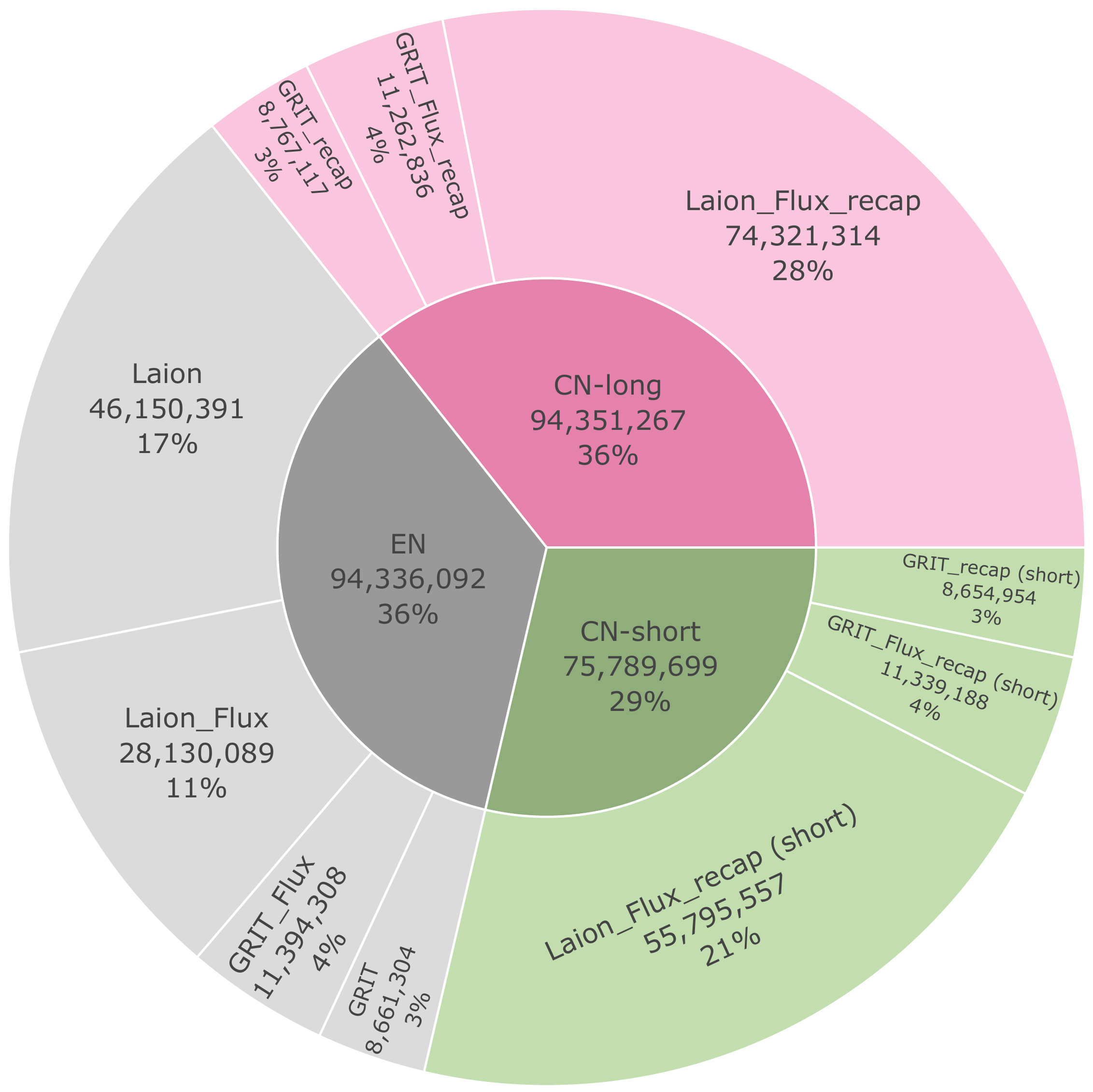} 
  \caption{Caption language distribution across different datasets used in Stage 1 training.
The root categories represent the caption language types: \textbf{EN} (English), \textbf{CN-long} (long-form Chinese), and \textbf{CN-short} (short-form Chinese).
Subcategories denote the data sources or processing steps: 
\textbf{GRIT} and \textbf{Laion} are public datasets where the captions were originally collected;
The suffix \textbf{Flux} indicates that the associated image was re-generated using Flux-dev;
The suffix \textbf{recap} denotes that the caption has been rewritten or adjusted during preprocessing.}

  \label{fig:dataset-language}
\end{figure}

In the first stage of training, we utilized a large-scale dataset consisting of mixed Chinese and English captions. The distribution of the "English", "Chinese-long", and "Chinese short" captions is illustrated in Figure~\ref{fig:dataset-language}. To ensure robust performance across captions of different lengths, we maintained a balanced proportion between short and long Chinese prompts during training. This design encourages the model to generalize well under diverse linguistic conditions, including semantic granularity and sentence complexities.

The construction of the dataset used for the first-stage training is illustrated in Figure~\ref{fig:dataset-language}. We collected approximately 150 million images, each paired with both Chinese and English captions, by using existing image-text datasets and generating additional captions using a recaptioning model. Specifically, we first collected image-caption pairs from the LAION~\cite{schuhmann2022laion5bopenlargescaledataset} and GRIT~\cite{fan2025gritteachingmllmsthink} datasets as the visual input for training, and supplemented the captions using CogVLM2~\cite{hong2024cogvlm2visuallanguagemodels} and Mantis~\cite{jiang2024mantisinterleavedmultiimageinstruction}. To further supplement the training set, we included images generated by Flux-dev~\cite{flux2024}, along with their corresponding captions produced by captioning models.

\subsection{Quantivate Evaluation}
\textbf{Evaluation on COCO.}~
Following prior work \cite{rombach2022high, podell2023sdxlimprovinglatentdiffusion, flux2024}, we evaluate CTA-Flux on the COCO \cite{lin2015microsoftcococommonobjects} 256×256 dataset using the zero-shot Fréchet Inception Distance (FID), a widely adopted metric for quantifying the quality and diversity of generated images.
We adopt a standardized benchmark based on the MS-COCO 30K dataset to ensure fair evaluation. Specifically, we use the Recaption-COCO~\cite{li2024recapdatacomp} dataset, reprocessing the original captions to produce more diverse and evenly distributed prompt lengths. Compared to the original COCO captions—typically short—the recaptioned prompts provide a more balanced length distribution, enabling more comprehensive and reliable assessment of model performance.
\begin{table}[t]
    \centering
    \caption{We compared CTA-FLux with other TTI models on the MS-COCO 256x256 dataset using FID-30K. The experimental setup included a classifier-free guidance scale of 3.5.}
   
        \begin{tabular}{@{}c c c c c c c c c c c@{}}
            \toprule
            Model & Input Language & FID $\downarrow$ \\
            \midrule
            LDM\cite{rombach2022high} & English & 37.45 \\
            SD\cite{rombach2022high} 1.5 & English & 22.87 \\
            SDXL\cite{podell2023sdxlimprovinglatentdiffusion} & English & 18.64 \\
            Flux\cite{flux2024}  & English & 16.39 \\
            
            \midrule
            BDM\cite{Liu_Cheng_Ma_Wu_Ma_Wu_Leng_Yin_2025} & Chinese and English & 28.34    \\
            CTA-Flux(Ours) & Chinese only &   \textbf{20.57}  \\
            CTA-FLux(Ours) & English only & \textbf{21.57} \\
            \bottomrule
        \end{tabular}

    \label{tab:coco}
\end{table}
As shown in Table~\ref{tab:coco}, although our model exhibits a slight increase in FID compared to Flux, the overall performance remains comparable. The experimental results demonstrate that our method successfully retains the English generation capability while introducing effective Chinese generation ability. Under fully Chinese-guided prompts, our model achieves an FID of 20.57, which is slightly worse than Flux but still significantly outperforms Stable Diffusion 1.5. Furthermore, our model achieves substantially better performance than BDM, while requiring only Chinese input, highlighting the efficiency and effectiveness of our multilingual adaptation approach.

\textbf{Chinese cultural tendency}~
To evaluate the cultural inclination of the model, we use GPT-4o to generate 100 culture-neutral prompts, consisting of 100 prompts in each of the four categories: people, architecture, cuisine, and festivals. We use these prompts to test the model and then compute the CLIP feature similarity between the generated images and the textual concepts "Chinese" and "Caucasian". The higher similarity indicates the model’s stronger inclination toward the corresponding cultural concept.

The experimental results, as shown in Table \ref{tab:clip-score-culture-irrelevant}, indicate that our CTA-FLux model exhibits a stronger cultural inclination towards Chinese culture than Flux. The similarity of CTA-FLux with "Caucasian" is nearly identical to Flux. Still, the similarity with "Chinese" is significantly higher than that of Flux, suggesting that during our training process, the model progressively acquired unique concepts of Chinese culture.

\begin{table}[t]
    \centering
    \caption{We test the CLIP score $\uparrow$ to measure the Chinese cultural inclination.}
    \resizebox{0.5\textwidth}{!}{
        \begin{tabular}{@{}c c c c c c c c c c c@{}}
            \toprule
            {{Class}   } & \multicolumn{2}{c}{{human}} & \multicolumn{2}{c}{{architecture}}  \\
              {Culture} & {Chinese} & {Caucasian} & {Chinese}  & {Caucasian}  \\
            \midrule
            Flux\cite{flux2024} & 2.00 & 4.18 & \textbf{3.67} & \textbf{5.83}   \\
            CTA-FLux(ours) & \textbf{2.34} & \textbf{5.22} & 3.15 & 5.32   \\

            \bottomrule
        \end{tabular}
    }
    \resizebox{0.5\textwidth}{!}{
        \begin{tabular}{@{}c c c c c c c c c c c@{}}
            \toprule
            {{Class}   } & \multicolumn{2}{c}{{food}} & \multicolumn{2}{c}{{festival}}  \\
               {Culture} & {Chinese} & {Caucasian} & {Chinese}  & {Caucasian}  \\
            \midrule
            Flux\cite{flux2024} & 4.11 & 3.28 & 8.64 & 6.85   \\
            CTA-FLux(ours) & \textbf{7.08} & \textbf{4.02} & \textbf{14.51} & \textbf{7.07}   \\

            \bottomrule
        \end{tabular}
    }
    
    \label{tab:clip-score-culture-irrelevant}
\end{table}

\subsection{Qualitative Results}
\textbf{Native language semantics.}

\begin{figure}[ht] 
  \centering
  \includegraphics[width=1.0\linewidth, ]{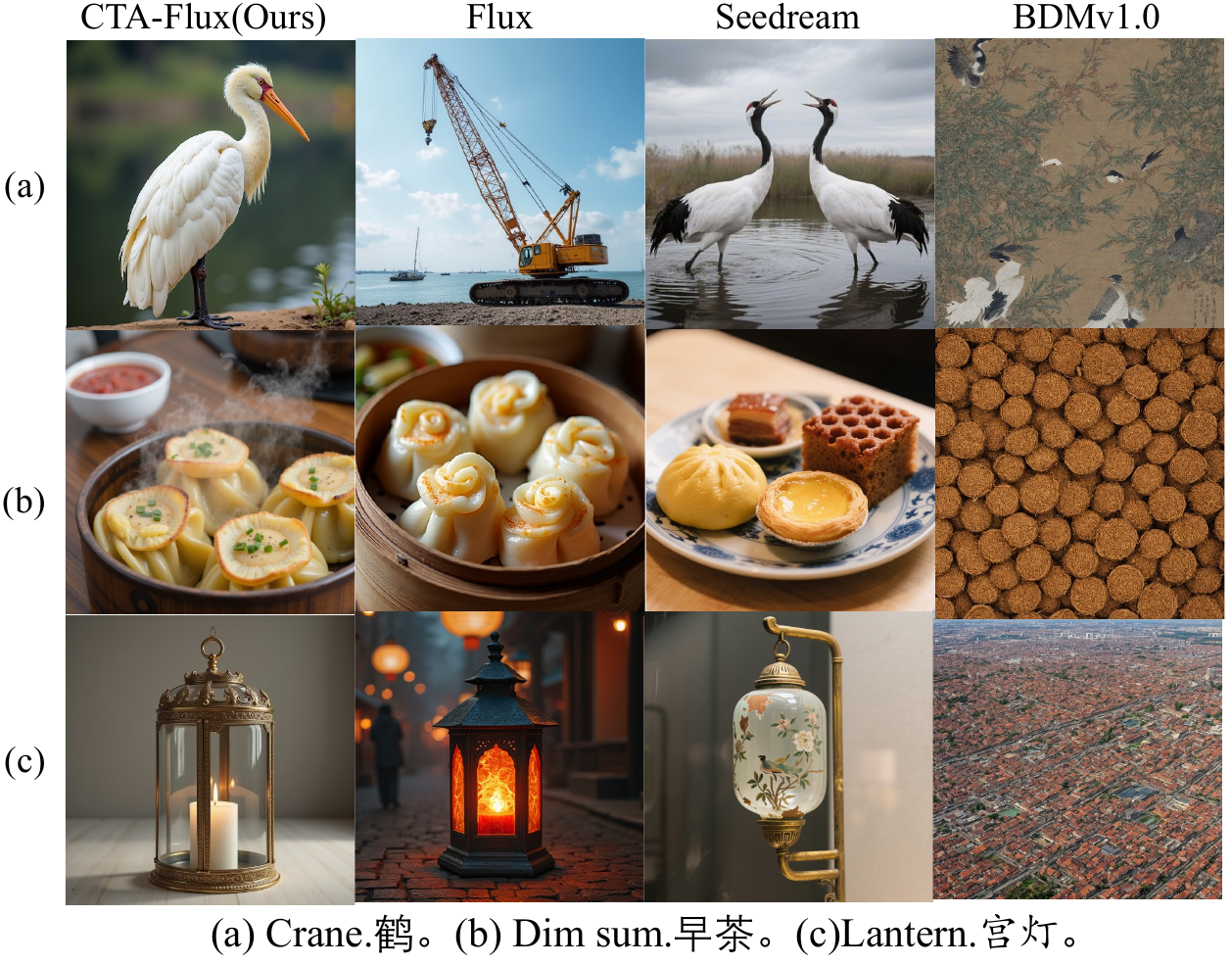}
  \caption{The performance of different TTI models in generating descriptions related to Chinese culture is compared. Our CTA-Flux not only maintains the generation quality of Flux but also exhibits a strong inclination towards Chinese culture and offers excellent community compatibility.
 }
 \vspace{-1em}
  \label{fig:native}
\end{figure}

We validated the Chinese semantics, and the experimental results are shown in Figure \ref{fig:native}. From the figure, it is evident that Flux~\cite{labs2025flux} faces the issue of ambiguity between Chinese and English. Additionally, Flux lacks cultural bias, whereas our CTA-Flux exhibits a stronger inclination towards Chinese culture. While BDMv1.0~\cite{Liu_Cheng_Ma_Wu_Ma_Wu_Leng_Yin_2025} supports Chinese prompt input and demonstrates a degree of cultural alignment with Chinese semantics, our CTA-Flux achieves superior performance in terms of image generation quality and aesthetic fidelity. Both our CTA-Flux and Seedream~\cite{gao2025seedream30technicalreport} demonstrate significant cultural bias, but compared to Seedream~\cite{gao2025seedream30technicalreport}, CTA-Flux is more compatible with various plugins in the open-source Flux community, offering better community support.

\begin{figure}[ht] 
  \centering
  \includegraphics[width=1.0\linewidth, height=0.86\linewidth]{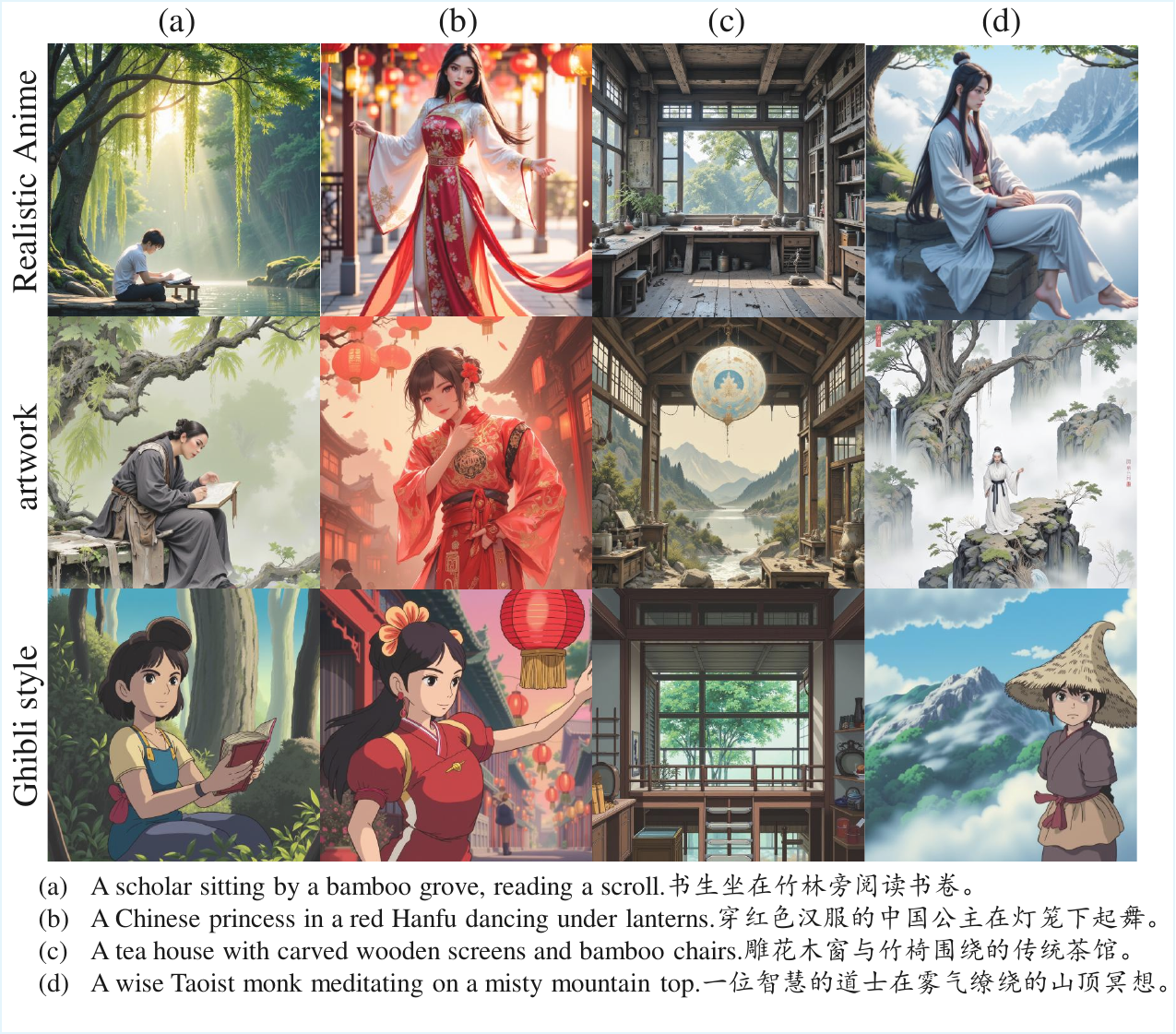}
  \caption{By using CTA-Flux with different LoRA modules, images generated with various LoRA prompts demonstrate CTA-Flux's strong community adaptability. This framework can generate high-quality, aesthetically pleasing, and high-resolution images.
 }
 \vspace{-1em}
  \label{fig:lora}
\end{figure}
\textbf{Community Support}~
To assess the extent of compatibility with the Flux community, we tested our model using several popular LoRA weights released by the community. The test results, as shown in Figure \ref{fig:lora}, demonstrate that our model is compatible with various styles of LoRA, exhibiting strong community.

\subsection{Ablation Study}

\textbf{Impact of Loss Function on Performance}~
To investigate the impact of adding loss and threshold on model performance, we conducted experiments under the same training conditions with 400,000 samples. We compared the model's generation quality using COCO FID \cite{heusel2018ganstrainedtimescaleupdate} and CLIP score \cite{radford2021learningtransferablevisualmodels} under different configurations: no loss, no threshold; no loss with threshold; and with loss but without threshold. The experimental results demonstrate in table~\ref{tab:loss} that adding loss improves the model's semantic understanding and generation quality. Furthermore, incorporating the threshold further enhances the generation quality while maintaining strong semantic understanding. 

\begin{table}[t]
    \centering
    \caption{We conducted tests with the same number of training steps for the ablation study on adding the loss function. The FID was evaluated using a fixed set of 3000 images selected from the COCO dataset as the test samples.}
    \begin{tabular}{@{}c|cccccc@{}}
        \toprule
        \multicolumn{4}{c}{Loss function} & \multicolumn{2}{c}{Metrics} \\
        \cmidrule(lr){1-4} \cmidrule(lr){5-6}
        \(\mathcal{L}_{RA}\) & \(\mathcal{L}_p\) & \(\mathcal{L}_{inter}\) & \(D_{threshold}\) & FID $\downarrow$ & CLIP score $\uparrow$ \\
        \midrule
        \xmark & - & -& -& 93.06  & 30.47 \\
        \multirow{3}{*}{\cmark} & \cmark & \xmark & \xmark & 77.10  & 31.42 \\
         & \cmark & \cmark & \xmark & 58.86  & 31.28 \\
         & \cmark & \cmark & \cmark & \textbf{53.52} & \textbf{31.40} \\
        \bottomrule
    \end{tabular}
    \label{tab:loss}
\end{table}

\textbf{Avoiding the Use of Chinese Embedding as Query in Cross-Attention
}~
In TTI generation models, text embeddings are typically used as queries in the cross-attention operation to facilitate information fusion across modalities. We conducted an ablation experiment to explore whether the Chinese text embeddings should be used as queries in the cross-attention computation and update their own features. We tested a model trained with 160,000 samples using COCO 256x256 FID and CLIP score metrics. The experimental results show that allowing the Chinese branch tokens to update during the cross-attention operation severely degrades the model's generation quality and semantic understanding. We speculate that this may be due to the strong random conditions introduced by the empty text input in the English branch during training, which, when the Chinese tokens query the English tokens, makes them highly susceptible to these random conditions, weakening the representation capability of the Chinese tokens.

\begin{table}[ht]
    \centering
    \caption{The ablation study on whether the Chinese embedding feature should be updated as a query was conducted.}

        \begin{tabular}{@{}c c c c c c c c c c c@{}}
            \toprule
            Chinese embedded token & FID  $\downarrow$ & CLIP score $\uparrow$ \\
            \midrule
           w/ update & 277.00  & 30.47   \\
           w/o update& \textbf{81.58}  & \textbf{31.42}  \\
            
            \bottomrule
        \end{tabular}

    \label{tab:update}
\end{table}

\section{Conclusion}

Our work realize the adaptation of the Flux model to Chinese text inputs by training a lightweight Chinese language branch within the MMDiT blocks, thereby avoiding the training of the entire TTI model from scratch. The English and image input branches remain frozen throughout, preserving compatibility with the base Flux model and the broader English TTI community. Through both qualitative visualizations and quantitative evaluations on multiple benchmarks, we demonstrate the effectiveness of our method in understanding Chinese text conditions and generating cultural and semantic-related contents. Moreover, our approach is not limited to Chinese-it is promising for broader multilingual TTI adaptation tasks, highlighting the generalization ability of the proposed framework.

\bibliography{aaai2026}

\begin{thebibliography}{46}
\providecommand{\natexlab}[1]{#1}

\bibitem[{Bai et~al.(2025)Bai, Chen, Liu, Wang, Ge, Song, Dang, Wang, Wang, Tang, Zhong, Zhu, Yang, Li, Wan, Wang, Ding, Fu, Xu, Ye, Zhang, Xie, Cheng, Zhang, Yang, Xu, and Lin}]{bai2025qwen25vltechnicalreport}
Bai, S.; Chen, K.; Liu, X.; Wang, J.; Ge, W.; Song, S.; Dang, K.; Wang, P.; Wang, S.; Tang, J.; Zhong, H.; Zhu, Y.; Yang, M.; Li, Z.; Wan, J.; Wang, P.; Ding, W.; Fu, Z.; Xu, Y.; Ye, J.; Zhang, X.; Xie, T.; Cheng, Z.; Zhang, H.; Yang, Z.; Xu, H.; and Lin, J. 2025.
\newblock Qwen2.5-VL Technical Report.
\newblock arXiv:2502.13923.

\bibitem[{Chen et~al.(2022)Chen, Liu, Zhang, Ye, Yang, and Wu}]{chen2022altclip}
Chen, Z.; Liu, G.; Zhang, B.-W.; Ye, F.; Yang, Q.; and Wu, L. 2022.
\newblock Altclip: Altering the language encoder in clip for extended language capabilities.
\newblock \emph{arXiv preprint arXiv:2211.06679}.

\bibitem[{Dhariwal and Nichol(2021)}]{dhariwal2021diffusion}
Dhariwal, P.; and Nichol, A. 2021.
\newblock Diffusion Models Beat GANs on Image Synthesis.
\newblock \emph{Advances in Neural Information Processing Systems}.

\bibitem[{Esser et~al.(2024)Esser, Kulal, Blattmann, Entezari, M{\"u}ller, Saini, Levi, Lorenz, Sauer, Boesel et~al.}]{esser2024scaling}
Esser, P.; Kulal, S.; Blattmann, A.; Entezari, R.; M{\"u}ller, J.; Saini, H.; Levi, Y.; Lorenz, D.; Sauer, A.; Boesel, F.; et~al. 2024.
\newblock Scaling rectified flow transformers for high-resolution image synthesis.
\newblock In \emph{Forty-first international conference on machine learning}.

\bibitem[{Evans(2010)}]{evans2010partial}
Evans, L.~C. 2010.
\newblock \emph{Partial Differential Equations}.
\newblock Providence, RI: American Mathematical Society.

\bibitem[{Fan et~al.(2025)Fan, He, Yang, Zheng, Kuo, Zheng, Narayanaraju, Guan, and Wang}]{fan2025gritteachingmllmsthink}
Fan, Y.; He, X.; Yang, D.; Zheng, K.; Kuo, C.-C.; Zheng, Y.; Narayanaraju, S.~J.; Guan, X.; and Wang, X.~E. 2025.
\newblock GRIT: Teaching MLLMs to Think with Images.
\newblock arXiv:2505.15879.

\bibitem[{Feng et~al.(2023{\natexlab{a}})Feng, Zhang, Yu, Fang, Li, Chen, Lu, Liu, Yin, Feng, Sun, Chen, Tian, Wu, and Wang}]{feng2023ernievilg20improvingtexttoimage}
Feng, Z.; Zhang, Z.; Yu, X.; Fang, Y.; Li, L.; Chen, X.; Lu, Y.; Liu, J.; Yin, W.; Feng, S.; Sun, Y.; Chen, L.; Tian, H.; Wu, H.; and Wang, H. 2023{\natexlab{a}}.
\newblock ERNIE-ViLG 2.0: Improving Text-to-Image Diffusion Model with Knowledge-Enhanced Mixture-of-Denoising-Experts.
\newblock arXiv:2210.15257.

\bibitem[{Feng et~al.(2023{\natexlab{b}})Feng, Zhang, Yu, Fang, Li, Chen, Lu, Liu, Yin, Feng et~al.}]{feng2023ernie}
Feng, Z.; Zhang, Z.; Yu, X.; Fang, Y.; Li, L.; Chen, X.; Lu, Y.; Liu, J.; Yin, W.; Feng, S.; et~al. 2023{\natexlab{b}}.
\newblock Ernie-vilg 2.0: Improving text-to-image diffusion model with knowledge-enhanced mixture-of-denoising-experts.
\newblock In \emph{Proceedings of the IEEE/CVF Conference on Computer Vision and Pattern Recognition}, 10135--10145.

\bibitem[{Gal et~al.(2022)Gal, Alaluf, Atzmon, Patashnik, Bermano, Chechik, and Cohen-Or}]{gal2022imageworthwordpersonalizing}
Gal, R.; Alaluf, Y.; Atzmon, Y.; Patashnik, O.; Bermano, A.~H.; Chechik, G.; and Cohen-Or, D. 2022.
\newblock An Image is Worth One Word: Personalizing Text-to-Image Generation using Textual Inversion.
\newblock arXiv:2208.01618.

\bibitem[{Gao et~al.(2025)Gao, Gong, Guo, Hou, Lai, Li, Li, Lian, Liao, Liu, Liu, Shi, Sun, Tian, Tian, Wang, Wang, Wang, Wang, Wang, Wu, Wu, Xia, Xiao, Zhai, Zhang, Zhang, Zhang, Zhao, Yang, and Huang}]{gao2025seedream30technicalreport}
Gao, Y.; Gong, L.; Guo, Q.; Hou, X.; Lai, Z.; Li, F.; Li, L.; Lian, X.; Liao, C.; Liu, L.; Liu, W.; Shi, Y.; Sun, S.; Tian, Y.; Tian, Z.; Wang, P.; Wang, R.; Wang, X.; Wang, X.; Wang, Y.; Wu, G.; Wu, J.; Xia, X.; Xiao, X.; Zhai, Z.; Zhang, X.; Zhang, Q.; Zhang, Y.; Zhao, S.; Yang, J.; and Huang, W. 2025.
\newblock Seedream 3.0 Technical Report.
\newblock arXiv:2504.11346.

\bibitem[{Gu et~al.(2022)Gu, Meng, Lu, Hou, Minzhe, Liang, Yao, Huang, Zhang, Jiang et~al.}]{gu2022wukong}
Gu, J.; Meng, X.; Lu, G.; Hou, L.; Minzhe, N.; Liang, X.; Yao, L.; Huang, R.; Zhang, W.; Jiang, X.; et~al. 2022.
\newblock Wukong: A 100 million large-scale chinese cross-modal pre-training benchmark.
\newblock \emph{Advances in Neural Information Processing Systems}, 35: 26418--26431.

\bibitem[{Heusel et~al.(2018)Heusel, Ramsauer, Unterthiner, Nessler, and Hochreiter}]{heusel2018ganstrainedtimescaleupdate}
Heusel, M.; Ramsauer, H.; Unterthiner, T.; Nessler, B.; and Hochreiter, S. 2018.
\newblock GANs Trained by a Two Time-Scale Update Rule Converge to a Local Nash Equilibrium.
\newblock arXiv:1706.08500.

\bibitem[{Ho, Jain, and Abbeel(2020)}]{ho2020denoising}
Ho, J.; Jain, A.; and Abbeel, P. 2020.
\newblock Denoising Diffusion Probabilistic Models.
\newblock \emph{Advances in Neural Information Processing Systems}.

\bibitem[{Hong et~al.(2024)Hong, Wang, Ding, Yu, Lv, Wang, Cheng, Huang, Ji, Xue, Zhao, Yang, Gu, Zhang, Feng, Yin, Wang, Qi, Song, Zhang, Liu, Xu, Li, Dong, and Tang}]{hong2024cogvlm2visuallanguagemodels}
Hong, W.; Wang, W.; Ding, M.; Yu, W.; Lv, Q.; Wang, Y.; Cheng, Y.; Huang, S.; Ji, J.; Xue, Z.; Zhao, L.; Yang, Z.; Gu, X.; Zhang, X.; Feng, G.; Yin, D.; Wang, Z.; Qi, J.; Song, X.; Zhang, P.; Liu, D.; Xu, B.; Li, J.; Dong, Y.; and Tang, J. 2024.
\newblock CogVLM2: Visual Language Models for Image and Video Understanding.
\newblock arXiv:2408.16500.

\bibitem[{Hu et~al.(2022)Hu, Shen, Wallis, Allen-Zhu, Li, Wang, Wang, Chen et~al.}]{hu2022lora}
Hu, E.~J.; Shen, Y.; Wallis, P.; Allen-Zhu, Z.; Li, Y.; Wang, S.; Wang, L.; Chen, W.; et~al. 2022.
\newblock Lora: Low-rank adaptation of large language models.
\newblock \emph{ICLR}, 1(2): 3.

\bibitem[{Jiang et~al.(2024)Jiang, He, Zeng, Wei, Ku, Liu, and Chen}]{jiang2024mantisinterleavedmultiimageinstruction}
Jiang, D.; He, X.; Zeng, H.; Wei, C.; Ku, M.; Liu, Q.; and Chen, W. 2024.
\newblock MANTIS: Interleaved Multi-Image Instruction Tuning.
\newblock arXiv:2405.01483.

\bibitem[{Kingma and Ba(2017)}]{kingma2017adammethodstochasticoptimization}
Kingma, D.~P.; and Ba, J. 2017.
\newblock Adam: A Method for Stochastic Optimization.
\newblock arXiv:1412.6980.

\bibitem[{Labs(2024)}]{flux2024}
Labs, B.~F. 2024.
\newblock FLUX.
\newblock \url{https://github.com/black-forest-labs/flux}.

\bibitem[{Labs et~al.(2025)Labs, Batifol, Blattmann, Boesel, Consul, Diagne, Dockhorn, English, English, Esser et~al.}]{labs2025flux}
Labs, B.~F.; Batifol, S.; Blattmann, A.; Boesel, F.; Consul, S.; Diagne, C.; Dockhorn, T.; English, J.; English, Z.; Esser, P.; et~al. 2025.
\newblock FLUX. 1 Kontext: Flow Matching for In-Context Image Generation and Editing in Latent Space.
\newblock \emph{arXiv preprint arXiv:2506.15742}.

\bibitem[{Li et~al.(2024)Li, Tu, Hui, Wang, Zhao, Xiao, Ren, Mei, Liu, Zheng, Zhou, and Xie}]{li2024recapdatacomp}
Li, X.; Tu, H.; Hui, M.; Wang, Z.; Zhao, B.; Xiao, J.; Ren, S.; Mei, J.; Liu, Q.; Zheng, H.; Zhou, Y.; and Xie, C. 2024.
\newblock What If We Recaption Billions of Web Images with LLaMA-3?
\newblock \emph{arXiv preprint arXiv:2406.12345}.

\bibitem[{Li et~al.(2023)Li, Chang, Rawls, Vulić, and Korhonen}]{Li_2023}
Li, Y.; Chang, C.-Y.; Rawls, S.; Vulić, I.; and Korhonen, A. 2023.
\newblock Translation-Enhanced Multilingual Text-to-Image Generation.
\newblock In \emph{Proceedings of the 61st Annual Meeting of the Association for Computational Linguistics (Volume 1: Long Papers)}, 9174–9193. Association for Computational Linguistics.

\bibitem[{Lin et~al.(2015)Lin, Maire, Belongie, Bourdev, Girshick, Hays, Perona, Ramanan, Zitnick, and Dollár}]{lin2015microsoftcococommonobjects}
Lin, T.-Y.; Maire, M.; Belongie, S.; Bourdev, L.; Girshick, R.; Hays, J.; Perona, P.; Ramanan, D.; Zitnick, C.~L.; and Dollár, P. 2015.
\newblock Microsoft COCO: Common Objects in Context.
\newblock arXiv:1405.0312.

\bibitem[{Lipman et~al.(2023)Lipman, Chen, Ben-Hamu, Nickel, and Le}]{lipman2023flow}
Lipman, Y.; Chen, R. T.~Q.; Ben-Hamu, H.; Nickel, M.; and Le, M. 2023.
\newblock Flow Matching for Generative Modeling.
\newblock In \emph{The Eleventh International Conference on Learning Representations}.

\bibitem[{Liu et~al.(2025)Liu, Cheng, Ma, Wu, Ma, Wu, Leng, and Yin}]{Liu_Cheng_Ma_Wu_Ma_Wu_Leng_Yin_2025}
Liu, S.; Cheng, B.; Ma, Y.; Wu, L.; Ma, A.; Wu, X.; Leng, D.; and Yin, Y. 2025.
\newblock Bridge Diffusion Model: Bridge Chinese Text-to-Image Diffusion Model with English Communities.
\newblock \emph{Proceedings of the AAAI Conference on Artificial Intelligence}, 39(5): 5541--5549.

\bibitem[{Liu, Gong, and Liu(2022)}]{liu2022flow}
Liu, X.; Gong, C.; and Liu, Q. 2022.
\newblock Flow straight and fast: Learning to generate and transfer data with rectified flow.
\newblock \emph{arXiv preprint arXiv:2209.03003}.

\bibitem[{Loshchilov and Hutter(2019)}]{loshchilov2019decoupledweightdecayregularization}
Loshchilov, I.; and Hutter, F. 2019.
\newblock Decoupled Weight Decay Regularization.
\newblock arXiv:1711.05101.

\bibitem[{Ni et~al.(2021)Ni, Ábrego, Constant, Ma, Hall, Cer, and Yang}]{ni2021sentencet5scalablesentenceencoders}
Ni, J.; Ábrego, G.~H.; Constant, N.; Ma, J.; Hall, K.~B.; Cer, D.; and Yang, Y. 2021.
\newblock Sentence-T5: Scalable Sentence Encoders from Pre-trained Text-to-Text Models.
\newblock arXiv:2108.08877.

\bibitem[{Podell et~al.(2023)Podell, English, Lacey, Blattmann, Dockhorn, Müller, Penna, and Rombach}]{podell2023sdxlimprovinglatentdiffusion}
Podell, D.; English, Z.; Lacey, K.; Blattmann, A.; Dockhorn, T.; Müller, J.; Penna, J.; and Rombach, R. 2023.
\newblock SDXL: Improving Latent Diffusion Models for High-Resolution Image Synthesis.
\newblock arXiv:2307.01952.

\bibitem[{Radford et~al.(2021)Radford, Kim, Hallacy, Ramesh, Goh, Agarwal, Sastry, Askell, Mishkin, Clark, Krueger, and Sutskever}]{radford2021learningtransferablevisualmodels}
Radford, A.; Kim, J.~W.; Hallacy, C.; Ramesh, A.; Goh, G.; Agarwal, S.; Sastry, G.; Askell, A.; Mishkin, P.; Clark, J.; Krueger, G.; and Sutskever, I. 2021.
\newblock Learning Transferable Visual Models From Natural Language Supervision.
\newblock arXiv:2103.00020.

\bibitem[{Ramesh et~al.(2021)Ramesh, Pavlov, Goh, Gray, Voss, Radford, Chen, and Sutskever}]{ramesh2021zero}
Ramesh, A.; Pavlov, M.; Goh, G.; Gray, S.; Voss, C.; Radford, A.; Chen, M.; and Sutskever, I. 2021.
\newblock Zero-shot text-to-image generation.
\newblock In \emph{International conference on machine learning}, 8821--8831. Pmlr.

\bibitem[{Rombach et~al.(2022)Rombach, Blattmann, Lorenz, Esser, and Ommer}]{rombach2022high}
Rombach, R.; Blattmann, A.; Lorenz, D.; Esser, P.; and Ommer, B. 2022.
\newblock High-Resolution Image Synthesis with Latent Diffusion Models.
\newblock \emph{arXiv preprint arXiv:2112.10752}.

\bibitem[{Saharia et~al.(2022)Saharia, Chan, Saxena, Li, Whang, Denton, Ghasemipour, Gontijo~Lopes, Karagol~Ayan, Salimans et~al.}]{saharia2022photorealistic}
Saharia, C.; Chan, W.; Saxena, S.; Li, L.; Whang, J.; Denton, E.~L.; Ghasemipour, K.; Gontijo~Lopes, R.; Karagol~Ayan, B.; Salimans, T.; et~al. 2022.
\newblock Photorealistic text-to-image diffusion models with deep language understanding.
\newblock \emph{Advances in neural information processing systems}, 35: 36479--36494.

\bibitem[{Schuhmann et~al.(2022)Schuhmann, Beaumont, Vencu, Gordon, Wightman, Cherti, Coombes, Katta, Mullis, Wortsman, Schramowski, Kundurthy, Crowson, Schmidt, Kaczmarczyk, and Jitsev}]{schuhmann2022laion5bopenlargescaledataset}
Schuhmann, C.; Beaumont, R.; Vencu, R.; Gordon, C.; Wightman, R.; Cherti, M.; Coombes, T.; Katta, A.; Mullis, C.; Wortsman, M.; Schramowski, P.; Kundurthy, S.; Crowson, K.; Schmidt, L.; Kaczmarczyk, R.; and Jitsev, J. 2022.
\newblock LAION-5B: An open large-scale dataset for training next generation image-text models.
\newblock arXiv:2210.08402.

\bibitem[{Sohl-Dickstein et~al.(2015)Sohl-Dickstein, Weiss, Maheswaranathan, and Ganguli}]{sohl2015deep}
Sohl-Dickstein, J.; Weiss, E.~A.; Maheswaranathan, N.; and Ganguli, S. 2015.
\newblock Deep Unsupervised Learning using Nonequilibrium Thermodynamics.
\newblock \emph{arXiv preprint arXiv:1503.03585}.

\bibitem[{Song, Meng, and Ermon(2020)}]{song2020denoising}
Song, J.; Meng, C.; and Ermon, S. 2020.
\newblock Denoising diffusion implicit models.
\newblock \emph{arXiv preprint arXiv:2010.02502}.

\bibitem[{Song and Ermon(2019)}]{song2019generative}
Song, Y.; and Ermon, S. 2019.
\newblock Generative Modeling by Estimating Gradients of the Data Distribution.
\newblock \emph{Advances in Neural Information Processing Systems}.

\bibitem[{Song, Meng, and Ermon(2023)}]{song2023consistency}
Song, Y.; Meng, C.; and Ermon, S. 2023.
\newblock Consistency Models.
\newblock \emph{arXiv preprint arXiv:2201.00367}.

\bibitem[{Song et~al.(2020)Song, Sohl-Dickstein, Kingma, Kumar, Ermon, and Poole}]{song2020score}
Song, Y.; Sohl-Dickstein, J.; Kingma, D.~P.; Kumar, A.; Ermon, S.; and Poole, B. 2020.
\newblock Score-Based Generative Modeling through Stochastic Differential Equations.
\newblock \emph{arXiv preprint arXiv:2011.13456}.

\bibitem[{Vaswani et~al.(2023)Vaswani, Shazeer, Parmar, Uszkoreit, Jones, Gomez, Kaiser, and Polosukhin}]{vaswani2023attentionneed}
Vaswani, A.; Shazeer, N.; Parmar, N.; Uszkoreit, J.; Jones, L.; Gomez, A.~N.; Kaiser, L.; and Polosukhin, I. 2023.
\newblock Attention Is All You Need.
\newblock arXiv:1706.03762.

\bibitem[{von Platen et~al.(2022)von Platen, Patil, Lozhkov, Cuenca, Lambert, Rasul, Davaadorj, Nair, Paul, Berman, Xu, Liu, and Wolf}]{von-platen-etal-2022-diffusers}
von Platen, P.; Patil, S.; Lozhkov, A.; Cuenca, P.; Lambert, N.; Rasul, K.; Davaadorj, M.; Nair, D.; Paul, S.; Berman, W.; Xu, Y.; Liu, S.; and Wolf, T. 2022.
\newblock Diffusers: State-of-the-art diffusion models.
\newblock \url{https://github.com/huggingface/diffusers}.

\bibitem[{Yang et~al.(2025)Yang, Li, Yang, Zhang, Hui, Zheng, Yu, Gao, Huang, Lv et~al.}]{yang2025qwen3}
Yang, A.; Li, A.; Yang, B.; Zhang, B.; Hui, B.; Zheng, B.; Yu, B.; Gao, C.; Huang, C.; Lv, C.; et~al. 2025.
\newblock Qwen3 technical report.
\newblock \emph{arXiv preprint arXiv:2505.09388}.

\bibitem[{Ye et~al.(2023)Ye, Zhang, Liu, Han, and Yang}]{ye2023ipadaptertextcompatibleimage}
Ye, H.; Zhang, J.; Liu, S.; Han, X.; and Yang, W. 2023.
\newblock IP-Adapter: Text Compatible Image Prompt Adapter for Text-to-Image Diffusion Models.
\newblock arXiv:2308.06721.

\bibitem[{Yin et~al.(2024)Yin, Gharbi, Zhang, Shechtman, Durand, Freeman, and Park}]{yin2024one}
Yin, T.; Gharbi, M.; Zhang, R.; Shechtman, E.; Durand, F.; Freeman, W.~T.; and Park, T. 2024.
\newblock One-step diffusion with distribution matching distillation.
\newblock In \emph{Proceedings of the IEEE/CVF conference on computer vision and pattern recognition}, 6613--6623.

\bibitem[{Yu et~al.(2022)Yu, Xu, Koh, Luong, Baid, Wang, Vasudevan, Ku, Yang, Ayan et~al.}]{yu2022scaling}
Yu, J.; Xu, Y.; Koh, J.~Y.; Luong, T.; Baid, G.; Wang, Z.; Vasudevan, V.; Ku, A.; Yang, Y.; Ayan, B.~K.; et~al. 2022.
\newblock Scaling autoregressive models for content-rich text-to-image generation.
\newblock \emph{arXiv preprint arXiv:2206.10789}, 2(3): 5.

\bibitem[{Zhang et~al.(2022)Zhang, Gan, Wang, Zhang, Zhang, Yang, Gao, Wu, Dong, He, Zhuo, Yang, Huang, Li, Wu, Lu, Zhu, Chen, Han, Pan, Wang, Wang, Wu, Zeng, and Chen}]{fengshenbang}
Zhang, J.; Gan, R.; Wang, J.; Zhang, Y.; Zhang, L.; Yang, P.; Gao, X.; Wu, Z.; Dong, X.; He, J.; Zhuo, J.; Yang, Q.; Huang, Y.; Li, X.; Wu, Y.; Lu, J.; Zhu, X.; Chen, W.; Han, T.; Pan, K.; Wang, R.; Wang, H.; Wu, X.; Zeng, Z.; and Chen, C. 2022.
\newblock Fengshenbang 1.0: Being the Foundation of Chinese Cognitive Intelligence.
\newblock \emph{CoRR}, abs/2209.02970.

\bibitem[{Zhang, Rao, and Agrawala(2023)}]{zhang2023adding}
Zhang, L.; Rao, A.; and Agrawala, M. 2023.
\newblock Adding conditional control to text-to-image diffusion models.
\newblock In \emph{Proceedings of the IEEE/CVF international conference on computer vision}, 3836--3847.

\end{thebibliography}

\end{document}